\documentclass[10pt,twocolumn,twoside,letterpaper]{IEEEtran}

\usepackage{geometry}
\makeatletter
\def\ps@IEEEtitlepagestyle{
  \def\@oddfoot{\mycopyrightnotice}
  \def\@evenfoot{}
}
\def\mycopyrightnotice{
  {\footnotesize
  \begin{minipage}{\textwidth}
  \centering
  978-1-7281-7693-2/20/\$31.00 \copyright2020 IEEE
  \end{minipage}
  }
}

\usepackage{cite}
\usepackage{amsmath,amssymb,amsfonts}
\usepackage{algorithmic}
\usepackage{graphicx,subfigure}
\usepackage{textcomp}
\usepackage{xcolor}
\usepackage{nopageno}

\usepackage{array}

\usepackage{url}

\begin{document}

\title{Deep Convolutional Neural Network for Low Projection SPECT Imaging Reconstruction}

\author{Charalambos Chrysostomou$^{*}$,
        Loizos Koutsantonis,
        Christos Lemesios,
        and Costas N. Papanicolas,
\thanks{C. Chrysostomou, L. Koutsantonis, C. Lemesios and C.N. Papanicolas are with the computation-based Science and Technology Research Center, The Cyprus Institute, 20 Konstantinou Kavafi Street, 2121, Aglantzia, Nicosia, Cyprus}

}

\maketitle
\let\thefootnote\relax\footnote{Manuscript received December 20, 2020}

\begin{abstract}
In this paper, we present a novel method for tomographic image reconstruction in SPECT imaging with a low number of projections. Deep convolutional neural networks (CNN) are employed in the new reconstruction method. Projection data from software phantoms were used to train the CNN network. For evaluation of the efficacy of the proposed method, software phantoms and hardware phantoms based on the FOV SPECT system were used. The resulting tomographic images are compared to those produced by the "Maximum Likelihood Expectation Maximisation" (MLEM).

\end{abstract}

\begin{IEEEkeywords}
Convolutional Neural Networks (CNN), Reconstruction, Single Photon Emission Computerized Tomography (SPECT), SPECT angle interpolation
\end{IEEEkeywords}

%
\IEEEpeerreviewmaketitle

\section{Introduction}

Computed tomography (CT) techniques are useful nuclear imaging techniques that are used most widely in medicine \cite{de2009projected}. Single Photon Emission Computerized Tomography (SPECT) \cite{wernick2004emission, madsen2007recent, mariani2010review} and Positron Emission Tomography (PET) \cite{cherry2001fundamentals, vaquero2015positron} are two of the most used CT techniques for diagnosis and monitoring of numerous diseases such as cancer \cite{perri2008octreo} and cardiac diseases \cite{chen2005onset}. SPECT image reconstructions have limited spatial resolution and underperform when a low number of projections is available or when the measurements are low quality and noisy. These difficulties produce an ill-posed inverse problem for image reconstruction. In order to solve those issues, alternative reconstruction algorithms were proposed and developed, such as ordered subset expectation maximization (OSEM) \cite{hudson1994accelerated} and maximization likelihood expectation maximization (MLEM) \cite{shepp1982maximum}. 
Nevertheless, the improvement in the reconstructions is not ideal in many cases. The reconstruction accuracy improves with an increase in the number of iterations but with the downside of also increasing the noise. The noise increase in the reconstruction is challenging to measure and may affect the final reconstruction as a false detection. Therefore, a new method is needed to provide high spatial resolution and handle high noise and a low number of projections. In this paper, a novel method for tomographic image reconstruction in SPECT imaging is proposed based on deep convolutional neural networks (CNNR).

The paper is organised as follows: Section \ref{sec:data} presents the data generated and used for training the proposed model, Section \ref{sec:cnn}, presents the proposed model. Section \ref{sec:results}, presents the results and discussions and finally Section \ref{sec:conclussion} is conclusions.

\section{Methods and Materials}
\subsection{Training Data}
\label{sec:data}

\begin{figure}[htp]
\centering
        \includegraphics[width=9cm]{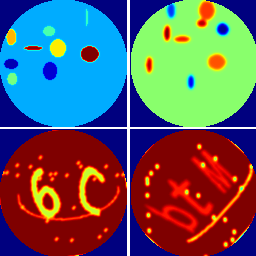}
\caption{Sample software phantoms randomly generated in order to train the proposed model}

\label{fig:random}

\end{figure}

A collection of sinograms was produced from randomly generated software phantoms through $Y_{i} = \sum_{j=1}^{N\,PxN\,R} P_{ij} F_{j}$ where $N\, R$ is the number of bin measurements per projection angle and $N\, P$ is the number of projection angles. For the proposed project, 500,000 software phantoms were generated on a rectangular grid of $128 \times 128$ pixels size. Furthermore, sinograms were created by simulating 24 projections, evenly spaced in 360 degrees, for the proposed model's training. Examples of the produced phantoms are presented in figure \ref{fig:random}. 

\begin{figure*}[!htp]
\centering
\subfigure[True activity of the Shepp Logan Phantom]{\centering \label{fig:shepp_logan_original}\includegraphics[width=60mm]{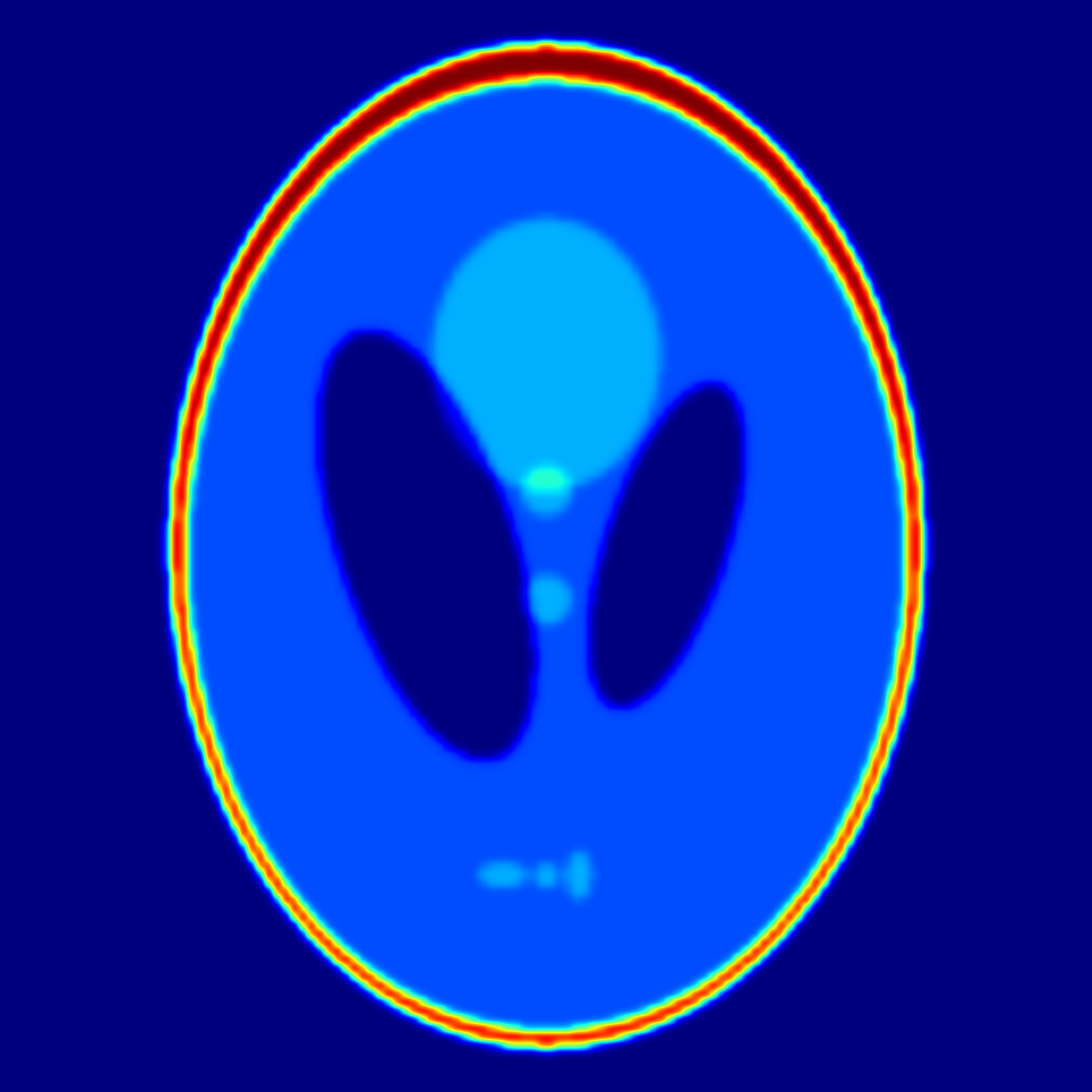}}
\subfigure[MLEM Reconstruction of the Shepp Logan Phantom]{\centering \label{fig:shepp_logan_mlem}\includegraphics[width=60mm]{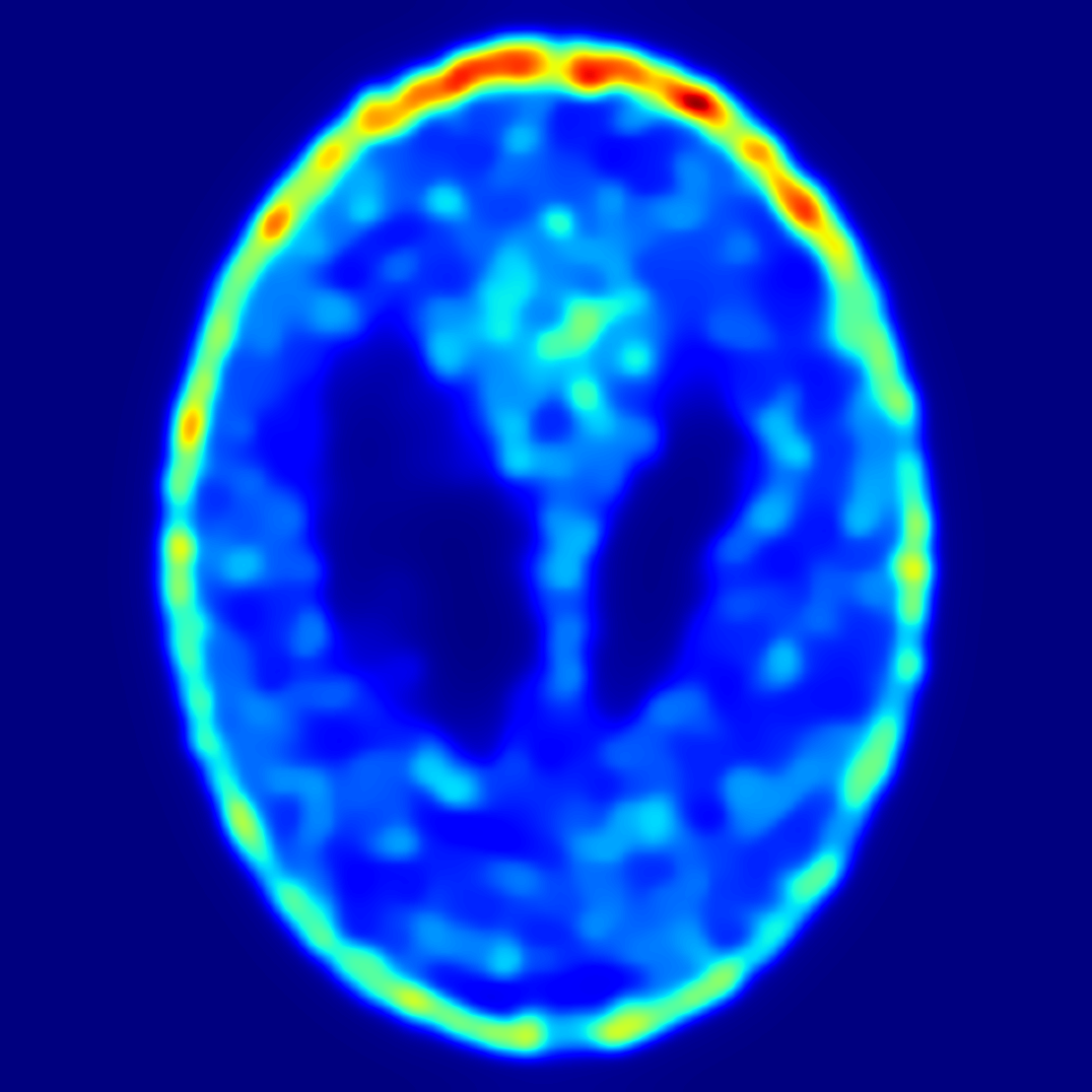}}
\subfigure[CNNR Reconstruction of the Shepp Logan Phantom] {\centering \label{fig:shepp_logan_cnnr}\includegraphics[width=60mm]{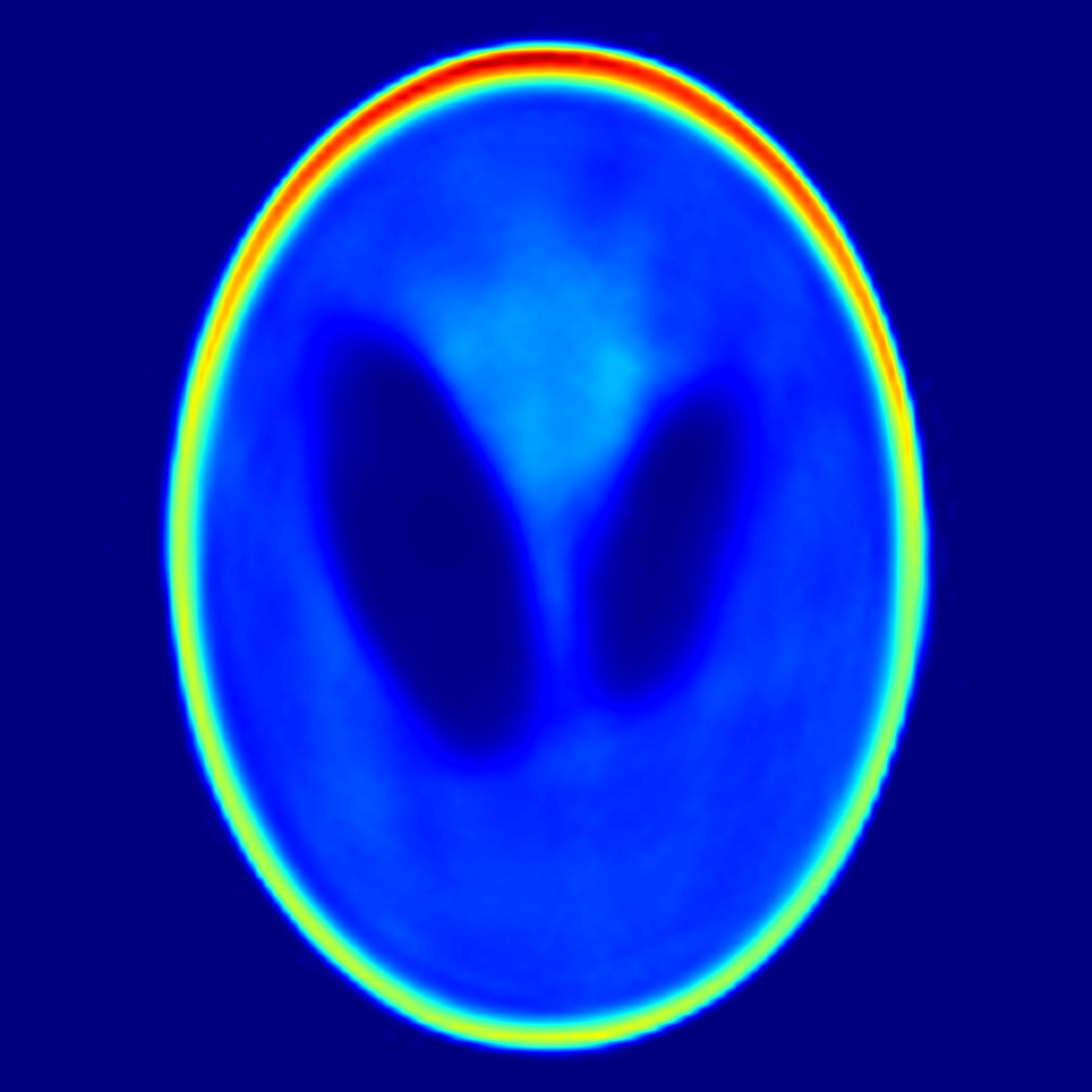}}
\caption{Evaluation of the proposed method results compared to the MLEM method using the Shepp Logan Phantom. The results achieved using the proposed method compare favourably to those obtained with the widely used MLEM method.}
\end{figure*}

In order to have more robust and accurate results from the proposed model, the produced sinograms were also randomised with a Poisson probability distribution to produce the noisy sets of projections. The model was trained for 1000 epochs, where 90\% of the available data was used for training and 10\% for validation. Ultimately, Shepp-Logan phantom \cite{shepp1974fourier}, as shown in figure \ref{fig:shepp_logan_original} was used to evaluate and assess the proposed method's capabilities.

\subsection{Proposed Model}
\label{sec:cnn}

\begin{figure*}[htp]
\centering
\includegraphics[width=18cm]{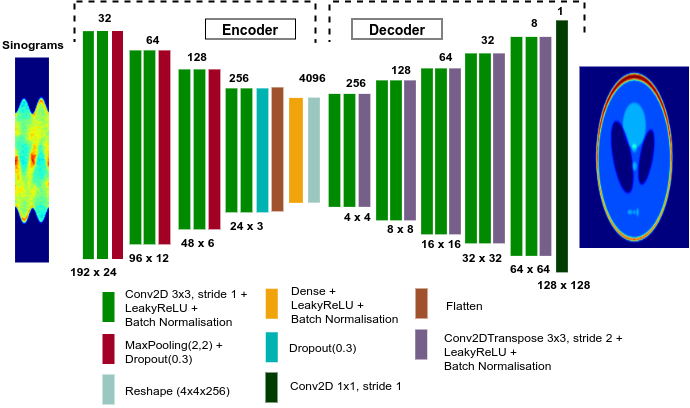}
  \caption{Structure of the proposed model based on deep convolutionl neural networks}
\label{fig:proposed_model}
\end{figure*}

Artificial Neural networks and, more specifically, Convolutional neural network (CNN) have been used with great success in multiple domains to analyse visual representations. Initially, CNN's were inspired by physiological and biological processes, and the connectivity model between neurons compares to the structure of the biological visual cortex \cite{hubel1962receptive}. CNN's are designed and require relatively minimum pre-processing of data compared to other methodologies by employing multilayer perceptrons \cite{schmidhuber2015deep}. This minimum prior knowledge provision provides an essential advantage of CNN's over other methodologies where prior knowledge and feedback from experts are required. CNN's have already successfully implemented in medical image analysis \cite{litjens2017survey}, and classification \cite{chrysostomou2018reconstruction, chrysostomou2019spect}. 

As shown in Figure \ref{fig:proposed_model}, the proposed model structure consists of two parts, the encoder and the decoder. The encoder consists of five blocks. The first three blocks consist of two convolutional layers with 3x3 kernel and rectified nonlinear activation function (Leaky ReLU) \cite{maas2013rectifier}, followed by a batch normalisation layer, and a 2x2 max pooling layer \cite{goodfellow2016deep} and a Dropout layer of 30\%. The fourth block consists of two convolutional layers with a 3x3 kernel, a batch normalisation layer, a Leaky ReLU activation function, a Dropout layer of 30\% and a flatten layer. The fifth layer consists of a dense layer of 4096 neurons, a Leaky ReLU activation function, and a batch normalisation layer. The number of kernels increases for each block, beginning with 32 kernels for the first block, 64, 128 and 256 for the second, third and fourth blocks, respectively.

Before the data are passed to the decoder, the decoder's output is transformed from 4096 features to a vector of 4x4x256. The decoder consists of six blocks. The first five blocks consist of two convolutional layers and a transposed convolutional layers, with a stride of 2,  with a 3x3 kernel and Leaky ReLU activation function and a batch normalisation layer. The final sixth layer consists of a convolutional layer of 1x1 kernel and linear activation function. The number of kernels decreases for each block, beginning with 256 kernels for the first block, 128, 64, 32, 8 and 1 for the second, third, fourth, fifth and sixth blocks, respectively. Structural Similarity (SSIM) Index \cite{wang2004image} was used as the loss function for the training of the proposed method, as shown in equation \ref{eq:SSMI}

\begin{equation}
  SSIM(x,y) = \frac{(2\mu_x\mu_y + C_1)  (2 \sigma _{xy} + C_2)} 
    {(\mu_x^2 + \mu_y^2+C_1) (\sigma_x^2 + \sigma_y^2+C_2)}
  \label{eq:SSMI}
\end{equation}

where $\mu _{x}$, $\mu _{y}$ and $\sigma _{x}^{2}$, $\sigma _{y}^{2}$ represent the average and variance of $x$ and $y$ respectively while $\sigma _{{xy}}$ represent the covariance of $x$ and $y$; $c_{1}=(k_{1}L)^{2}$, $c_{2}=(k_{2}L)^{2}$ two variables to stabilise the division with weak denominator; $L$ the dynamic range of the pixel-values (typically this is $2^{{\#bits\ per\ pixel}}-1)$; $k_{1}=0.01$ and $k_{2}=0.03$ by default.

\section{Results and Discussions}
\label{sec:results}

\addtolength{\textheight}{-1.3cm}

\begin{figure}[!htp]
    \centering
        \includegraphics[width=7cm]{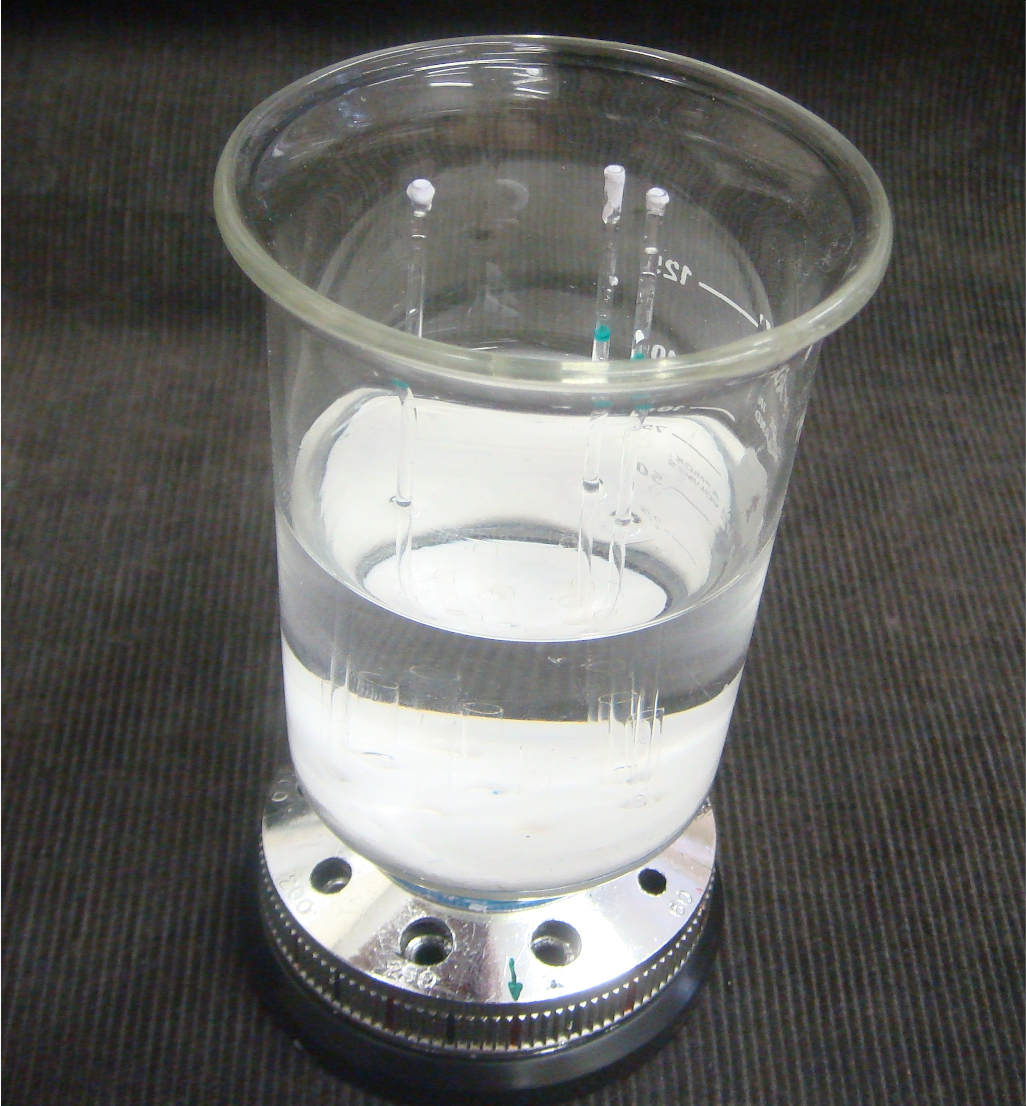}
    \caption{Hardware phantoms of three parallel pillars, with activity of 21 $\mu C i$ and variant background activity of 190, 360 and 700 $\mu C i$}
\label{fig:experiment}
\end{figure}

\begin{figure*}[!htp]
    \centering
        \includegraphics[width=18cm]{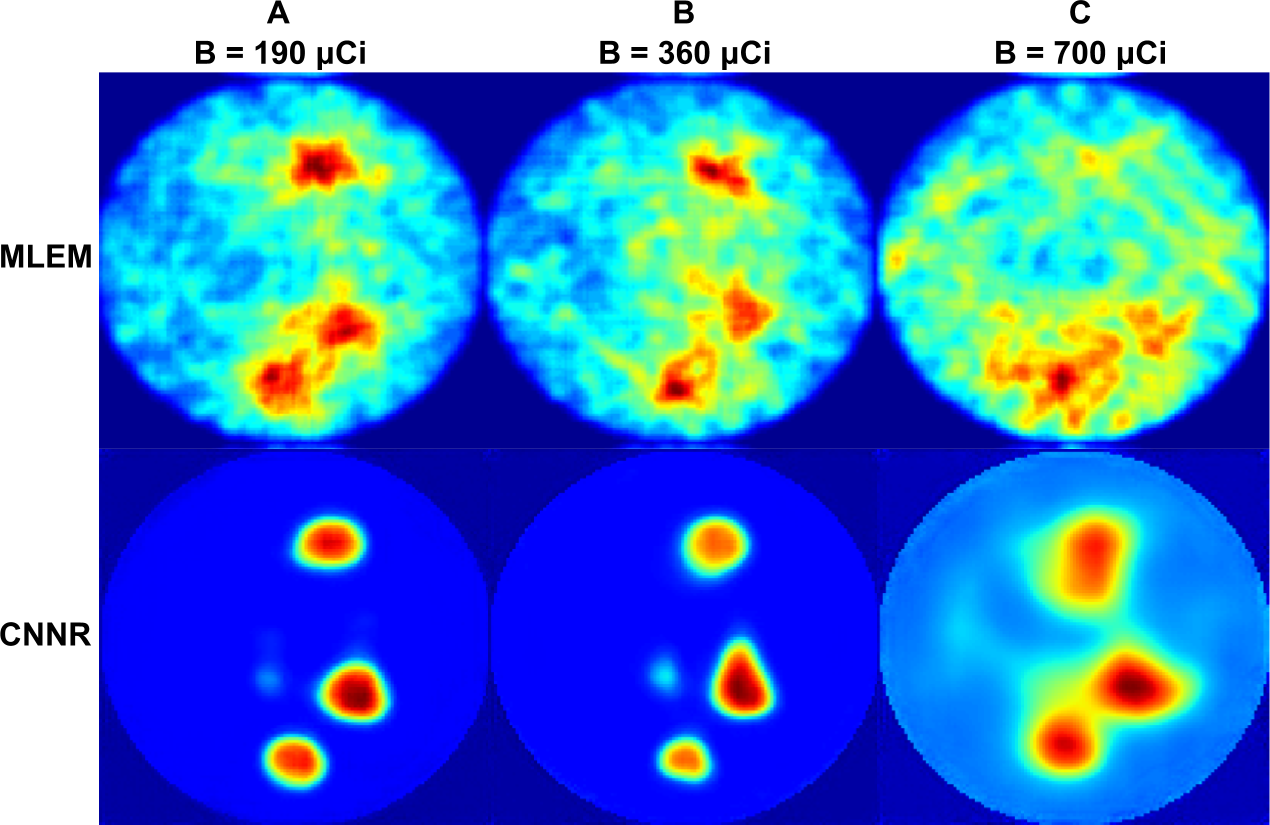}
    \caption{Evaluation of the proposed method results compared to the MLEM method using the Hardware phantoms.}
\label{fig:hw_results}
\end{figure*}

The performance of the proposed method was compared against existing methods by using the Mean Square Error (MSE), Structural Similarity (SSIM) Index \cite{wang2004image}, and the Pearson Correlation Coefficient (PCC) \cite{benesty2009pearson}. As the results show and illustrated in Table \ref{table:results}, the proposed methodology outperforms MLEM with 0.003 versus 0.006, 0.92 versus 0.77 and 0.96 versus 0.93 for MSE, SSIM and PCC respectively. The proposed method was also tested using hardware phantoms generated using a small FOV SPECT system \cite{spanoudaki2004design} of three parallel pillars, as shown in Figure \ref{fig:experiment} with the activity of 21 $\mu C i$ and variant background activity of 190, 360 and 700 $\mu C i$. The results are presented in figure \ref{fig:hw_results}.

\begin{table}[htp]
\renewcommand{\arraystretch}{1.5}
\centering
\caption{Results}
\label{table:results}
\begin{tabular}{c  c  c  c }
\hline
Methodology & MSE & SSIM & PCC \\
 \hline
MLEM & 0.006 & 0.77 & 0.93\\
Proposed Model  & 0.003 & 0.92 & 0.96\\

\hline
\end{tabular}
\end{table}

\section{Conclusions}
\label{sec:conclussion}

In this paper, we proposed a new method based on convolutional neural networks to reconstruct SPECT imaging and demonstrate its capability to reconstruct images with a limited number of projections. As the results show, the proposed method can significantly outperform existing methods such as MLEM. Although the results presented in this proposed work are appropriate for representing the proposed method's capabilities, further experimentation is needed, as shown in Figure \ref{fig:experiment} were the size and shape of the reconstructed caterpillars where overestimated.

\bibliography{ref} 
\bibliographystyle{IEEEtran}

\end{document}